\documentclass{article}

\PassOptionsToPackage{numbers, comma, compress, sort}{natbib}

\usepackage[preprint]{neurips_2024}

\usepackage[utf8]{inputenc} 
\usepackage[T1]{fontenc}    
\usepackage{hyperref}       
\usepackage{url}            
\usepackage{booktabs}       
\usepackage{amsfonts}       
\usepackage{nicefrac}       
\usepackage{microtype}      
\usepackage{xcolor}         
\usepackage{graphicx}
\usepackage{arydshln}
\usepackage{natbib}

\newcommand{\specialcell}[2][c]{%
  \begin{tabular}[#1]{@{}c@{}}#2\end{tabular}}

\title{Uncertainty and Generalizability in Foundation Models for Earth Observation}

\author{%
  Raúl Ramos-Pollán \\
  Universidad de Antioquia, Colombia\\
  \texttt{raul.ramos@udea.edu.co}
  \And
  Freddie Kalaitzis \\
  University of Oxford, UK\\
  \texttt{freddie.kalaitzis@cs.ox.ac.uk} 
  \And
  Karthick Panner Selvam\\
  University of Luxembourg\\
  \texttt{karthick.pannerselvam@uni.lu} 
  }

\begin{document}

\maketitle

\begin{abstract}
We take the perspective in which we want to design a downstream task (such as estimating vegetation coverage) on a certain area of interest (AOI) with a limited labeling budget. By leveraging an existing Foundation Model (FM) we must decide whether we train a downstream model on a different but label-rich AOI hoping it generalizes to our AOI, or we split labels in our AOI for training and validating. In either case, we face choices concerning what FM to use, how to sample our AOI for labeling, etc. which affect both the performance and uncertainty of the results. In this work, we perform a large ablative study using eight existing FMs on either Sentinel 1 or Sentinel 2 as input data, and the classes from the ESA World Cover product as downstream tasks across eleven AOIs. We do repeated sampling and training, resulting in an ablation of some 500K simple linear regression models. Our results show both the limits of spatial generalizability across AOIs and the power of FMs where we are able to get over 0.9 correlation coefficient between predictions and targets on different chip level predictive tasks. 
And still, performance and uncertainty vary greatly across AOIs, tasks and FMs. We believe this is a key issue in practice, because there are many design decisions behind each FM and downstream task (input modalities,  sampling, architectures, pretraining, etc.) and usually a downstream task designer is aware of and can decide upon a few of them. Through this work, we advocate for the usage of the methodology herein described (large ablations on reference global labels and simple probes), both when publishing new FMs, and to make informed decisions when designing downstream tasks to use them.



\end{abstract}

\section{Introduction}
Foundation Models (FMs) have been proven particularly useful in scarce label scenarios, where models are built with a limited amount of labeled data for different downstream tasks. Earth Observation (EO) is not an exception, and EO labeled data with its own particularities. It is usually quite heterogeneous, being available in the particular regions where the labelling effort is focused for each specific downstream task. Furthermore, there are more abundant in regions like the US and Europe where science resources are larger.

In this work, we assume we have a limited labeling budget on the AOI for the downstream task of interest, the \textbf{target AOI} and we want to make the best use of it by using an FM. We consider two cases: (1) there is an \textbf{external AOI} having a healthy amount of labels to train a downstream model for our task, therefore we would only need to use our label budget in the target AOI for validation; and (2) we split our labeling effort in our target AOI both for training a model and validating it.

We are firstly concerned with \textbf{spatial generalizability}, dealing with models trained in one AOI doing inference somewhere else. Performance can be compromised due to many reasons, because land features are simply different across AOIs (vegetation, buildings, etc.), because the FM did not capture enough information, because we are not training with sufficient data, etc.  Secondly we are concerned with the \textbf{uncertainty} stemming from small data scenarios (epistemic uncertainty, see \cite{hullermeier2021aleatoric}).

We test eight different FMs using Sentinel 1 (S1) amplitude and Sentinel 2 (S2) as input data to obtain chip level embeddings. Then, we use the ESA World Cover product (\texttt{esawc}) as global labels for a chip level percentage estimation downstream task using linear regression for each dataset class (tree cover, built up, permanent water, etc.) so that we can perform ablations and tests across the globe. With this we hope to provide a first indication on what to expect when applying FMs to related downstream tasks (for instance biomass estimation, human footprint, etc.)

We use a simple linear regression (LR) model to make a large ablation study (over 500K models trained) using the US, Europe and China as external AOIs and Colombia, Perú, India, Kenya, California, Texas, Spain and Germany as target AOIs. LR provides us both computational affordability for such a large study and a glimpse on the straight forward information content provided by FM embeddings.

The results below show a wide diversity of situations. Generalizability is higher in some downstream tasks and AOIs than in others; as we increase the labeling effort in target AOIs we reduce the uncertainty of our experiments on different degrees for different FMs and tasks; different sampling methods provide different outcomes, etc. But in general we see that all FMs behave similarly, with certain advantages of one FM over the other for certain specific situations. And yet, it is surprising the level of predictive power we can obtain in several cases just using a linear probe directly on FM embeddings.

In all, there are many design decisions behind each FM and downstream task (input modalities, data sampling, architectures, pretraining method, etc.) and a downstream task designer is usually aware and can decide upon a few of them. Through this work, we advocate for the usage of the methodology herein described (reference global labels and simple probes), both when publishing new FMs, and to make informed decisions when designing downstream tasks to use them.

\section{Previous works}
\label{sec:prevworks}
During the last few years there has been a frentic effort to develop a wide range of Earth Observation Foundation Models, covering different input sensors, formats, time series, world regions, etc. See for instance \cite{jakubik2023foundationmodelsgeneralistgeospatial} \cite{cong2022satmae}  \cite{stewart2023ssl4eo} \cite{guo2024skysense} \cite{smith2024earthpttimeseriesfoundation}  \cite{allen2023largescalemaskedautoencoding} \cite{allen2023fewshotlearningglobalmultimodal} \cite{10490262} just to name a few. Vision Transformers (ViT) \cite{dosovitskiy2021image} as architecture and masked autoencoders based losses \cite{he_masked_2021} dominate this landscape, and most FMs draw insights and methods from the self supervised community \cite{wang2022self}.

Several reviews pinpoint many aspects and challenges for FMs in Earth Observation \cite{lu2024aifoundationmodelsremote} \cite{mai2023opportunities}. Key aspects among those challenges are \textbf{engineering} (how easy it is to inject my data into FMs), \textbf{semantics} (how sensible are FMs to the earth features I am interested in ), \textbf{multimodality} (how well can the FM exploit inputs from multiple sensors), \textbf{flexibility} (how robust are FMs to missing data and different time and spatial resolutions), \textbf{generalizability} (can I use FMs to build models doing inference across regions?) and \textbf{uncertainty} (how much variance should I expect when feeding small sampled data to FMs?). In this work we address the last two.

Finally, some efforts have focused on creating benchmarks or benchmarking datasets for EO. See for instance \cite{fibaek2024phileobenchevaluatinggeospatial} \cite{lacoste2023geobenchfoundationmodelsearth}. The focus of this work is complementary, since we are trying to obtain a comprehensive view of FMs generalizability and uncertainty rather than focusing on complex end-to-end downstream tasks.

\section{Foundation models, data and downstream tasks}
\label{sec:data}

\begin{figure}[ht]
\caption{Distributions of the seven selected \texttt{esawc} classes for this work on each AOI. Percentage number indicate how much of the AOI landmass is covered by those seven classes. Dotted line separates external AOIs from target AOIs. }
\centering
\includegraphics[width=1.0\textwidth]{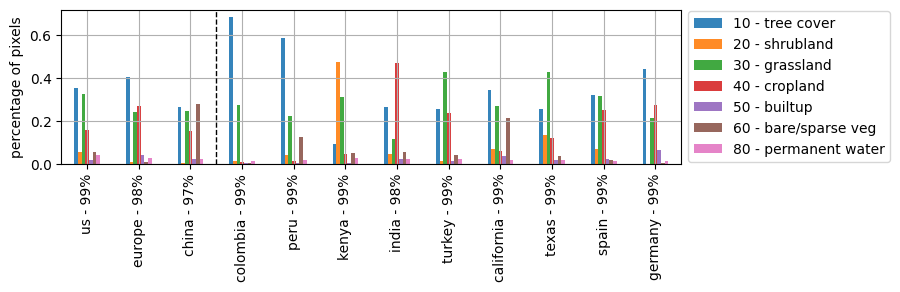}
\label{fig:esawc-distributions}
\end{figure}

\paragraph{Areas of Interest (AOIs)} Table \ref{tab:aois} details the AOIs used in this work. We selected three \textbf{external AOIs} (US, Europe, China) typically rich on labels upon which we train models; and test them in the \textbf{target AOIs} (Colombia, Peru, Kenya, India, Spain, Germany, California, Texas). They were selected to intuitively represent a diversity of terrain features, which can be observed in Figure \ref{fig:esawc-distributions}. The inclusion of European countries and US states in the target AOIs is to have a sense on how region wide models perform on included regions.

\begin{table}[]
\begin{center}
\begin{tabular}{lcccc}
\hline
\textbf{AOI} & \textbf{usage} & \textbf{Mkm$^2$} & \textbf{number of chips} & \textbf{landmass coverage} \\ \hline\hline
US & external &  7.9 &  30418 & 10\%   \\
Europe & external & 5.5 & 35936 & 17\%   \\
China & external & 9.4 & 35934 & 10\%   \\ \hline
Colombia & target & 1.2 & 4222 & 10\%   \\
Peru & target & 1.3 & 4860 &  10\%   \\
India & target & 3.1 & 11875 & 10\%   \\
Kenya & target & 0.6 & 2228 & 10\%  \\ 
Spain & target & 0.5 & 1905 & 10\%   \\
Germany & target & 0.5 & 6556 & 48\%   \\
California & target & 0.4 & 1589 & 10\%  \\
Texas & target & 0.7 & 2620 & 10\%   \\
\hline
\end{tabular}
\caption{Areas of Interest (AOIs) used in this work. \textbf{External} AOIs are only used for training downstream models. \textbf{Target} AOIs are used both for training and validation. We sampled randomly 10\% of the world landmass, except in Europe where we sampled 100\% on an area of 1000km centered in Luxembourg.
\label{tab:aois} 
}
\end{center}
\end{table}

\paragraph{Input data to FMs} We use chips of size 
$512\times 512$ pixels which, at Sentinels resolution of 10m/pixels, corresponds to a spatial resolution of
 5.12km$\times$5.12km, or 26.2km$^2$. We created S1 and S2 datasets using 
 \texttt{geetiles}\footnote{\url{http://github.com/rramosp/geetiles}} taking data for a full year and computing 
 the median per pixel per season (winter, spring, summer, fall). For S1 this results in 16 channels (vv + vh, ascending/descending per season), and for S2 this results in 44 channels (11 bands per session). The S2 bands used are [B2, B3, B4, B5, B6, B7, B8, B8A, B11, B12] as accepted by all S2 models.
  
\paragraph{Foundation Models} We used eight foundation models as detailed in  Table \ref{tab:fms}. They were selected using criteria of easiness to use, similar model size and pretrained on Sentinel imagery. Each FM either accepts S1 or S2 chips as input data and produces an embeddings vector for each chip. This embeddings vector is then used as input to the linear models detailed below. So, for each of our AOIs there a set of embeddings vectors for all its chips for each FM. Note that different FMs are trained on different world regions and/or sampling methods. 

\begin{table}[]
\begin{center}
\begin{tabular}{lcccccc}
\hline\hline
Foundation Model & \specialcell[c]{emb. \\ size} & 
                   params& 
                   arch. & 
                   \specialcell[c]{pretrain\\method} & 
                   \specialcell[c]{AOI used to\\train the FM}  \\ \hline
\hline
\textbf{Sentinel 1} models\\
\hline
\texttt{s1-fdl2024mae} \cite{allen2023largescalemaskedautoencoding} $^{\star}$  &  768 & 95M & ViT & MAE & world 10\% random \\
\texttt{s1-fdl2024clip} \cite{allen2023fewshotlearningglobalmultimodal}  $^{\star}$ & 768 & 90M & ViT &  CLIP & world 10\% random \\
\texttt{s1-clay\_v02} $^{\dagger \natural}$ & 768 & 113M & ViT & MAE & world landcover stratified\\
\texttt{s1-clay\_v1} $^{\dagger \natural}$ & 768  & 201M  & ViT & MAE & world landcover stratified \\
\hline
\textbf{Sentinel 2} models \\
\hline
\texttt{s2-fdl2024mae} \cite{allen2023largescalemaskedautoencoding}  $^{\star}$ & 768 & 98M & ViT & MAE & world 10\% random\\
\texttt{s2-fdl2024clip} \cite{allen2023largescalemaskedautoencoding} $^{\star}$ & 768 & 91M & ViT & CLIP& world 10\% random\\
\texttt{s2-clay\_v02} $^{\dagger \natural}$ & 768 & 113M & ViT & MAE & world landcover stratified\\
\texttt{s2-prithvi} \cite{jakubik2023foundationmodelsgeneralistgeospatial} $^{\ddagger}$ &  768 & 116M & ViT & MAE & US climate stratified\\
\hline\hline
\end{tabular}
\caption{Foundation Models (FMs) used in this work. FMs with $\star$ take as input separated median per pixel per season (8 channels for vv, vh for seasons for S1, 11 channels for S2) to produce a single embeddings vector. FMs with $\dagger$ are feed sequentially each season chip, producing four embeddings vectors which are then averaged into a single one. Both S1 and S2 CLIP encoders were trained together. Prithvi ($\ddagger$) accepts three time steps so we input the first three season medians (winter, spring, summer). \texttt{clay\_v1} also accepts S2 data but was left out because of timing constraints. Clay models (${\natural}$) can be found at \url{https://github.com/Clay-foundation/model}
\label{tab:fms} 
}
\end{center}

\end{table}

\paragraph{Downstream tasks} We use the ESA World Cover product \cite{esawc} (\texttt{esawc}) which delivers 11 landcover classes at the same spatial resolution as S1 and S2 (10m/pixel).  Out of those we focus only on the seven classes in Figure \ref{fig:esawc-distributions}, since the rest occur very rarely, and the figure shows their distribution across the selected AOIs. For each of those classes, we set up a chip-level regression task to predict the percentage of pixels in the chip labeled with that class. This is a number between 0 and 1 per chip.

\section{Methods}
\label{sec:method}

\paragraph{Linear regression on chip level percentage} We train a linear regression model for each experimental
setup described below (\texttt{esawc} class, train AOI, target AOI, etc.)  
 By just using linear regression we intend to have a sense on the \emph{raw} information content provided by FMs' embeddings with respect to the different \texttt{esawc} classes. Also, with the large number of experiment combinations due to the ablations shown below, linear regression seems a reasonable choice to make this work computationally feasible.

\begin{figure}[ht]
\caption{Example predictions of a linear probe on the embedding from FM \texttt{s1-fdl2024-mae}, trained with data from Europe on different \texttt{esawc} tasks and target AOIs. An asterisk \texttt{[*]} denotes a correlation coefficient greater than 0.7 as the threshold above which we will consider the embeddings do contain useful information  for that task and target AOI. }
\centering
\includegraphics[width=1.0\textwidth]{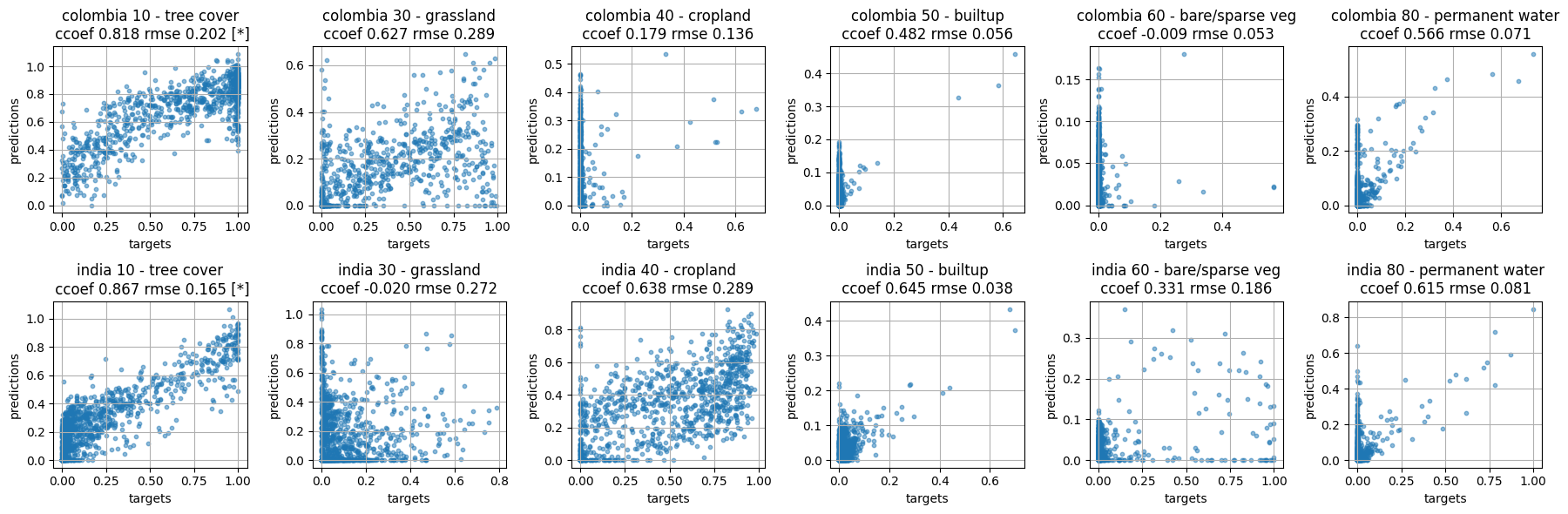}
\label{fig:example-predictions}
\end{figure}

\paragraph{Metric}
We use the correlation coefficient between predictions and targets. Observe how, in Figure \ref{fig:example-predictions} the correlation coefficient seems to capture better than RMSE whether the linear probe on the embeddings is actually able to perform the task. Notice, for instance, Colombia on class \texttt{permanent water} with misleading low RMSE, or India on \texttt{tree cover}  vs. \texttt{bare/sparse vegetation} with similar RMSE, but the model clearly picking up the first class, but not the second. For interpretation of the results we establish a threshold of 0.7 above which we consider FM embeddings are actually picking up meaningful information on the underlying predictive task.

\paragraph{Sampling}
We use four different sampling methods when selecting data for training, both when training with the external AOIs and the target AOIs. This sampling is done on the datasets described in Table \ref{tab:aois}. With \textbf{esawc sampling} we sample proportional to distribution of \texttt{esawc} classes in the AOI, so that the presence of all classes is as uniform as possible. With \textbf{fps sampling} we use Furthest Point Sampling \cite{qi2017pointnet++} on the embeddings for each FM using euclidean distance, which intuitively favors an overall variance of the embeddings. With \textbf{random sampling} we do a spatially uniform random sampling. And with \textbf{srtm sapling} we do sampling proportional to the mean elevation of each chip so that we get as many different elevations as possible.

\paragraph{Experimentation}
We did two sets of ablations (1) using external AOIs for training models, and testing on target AOIs; and (2) using part of the data in the target AOIs for training, and testing on the remaining target AOI data. We ablated on the following hyperparameters: FM embeddings (see Table \ref{tab:fms}), external and target AOIs (see Table \ref{tab:aois}), number of train elements on external AOI [300,3000,30000], number of train elements on the target AOI [10,50,100,500], number of test elements on the target AOI [10,50,100,500] and \texttt{esawc} class (7 classes).

This results in  $\sim$18K models when using external AOIs for training, and $\sim$7K models when splitting target AOIs for train and test. In this later case, we don't do cross AOI training and testing. We repeat each experiment 20 times, resampling train and test data each time and reporting the mean and standard deviation of the metric. Sampling test data is always done using spatially uniform random sampling. With this, in total we trained some 500K linear regression models.

\section{Results}
\label{sec:results}

\paragraph{Foundation Models on different target AOIs and tasks}

\begin{figure}[h]
\caption{Overall view of linear probes with different train AOIs, target AOIs and dowstrean tasks, showing models trained in 30K elements in external AOIs and tested with 500 elements in target AOIs. Showing the mean of 20 runs. Correlation threshold is set at 0.7 (white). Bluer positions represent greater correlation between predictions and targets, redder ones worse. Black horizontal line splits S1 and S2 FMs.}
\centering
\includegraphics[width=1.0\textwidth]{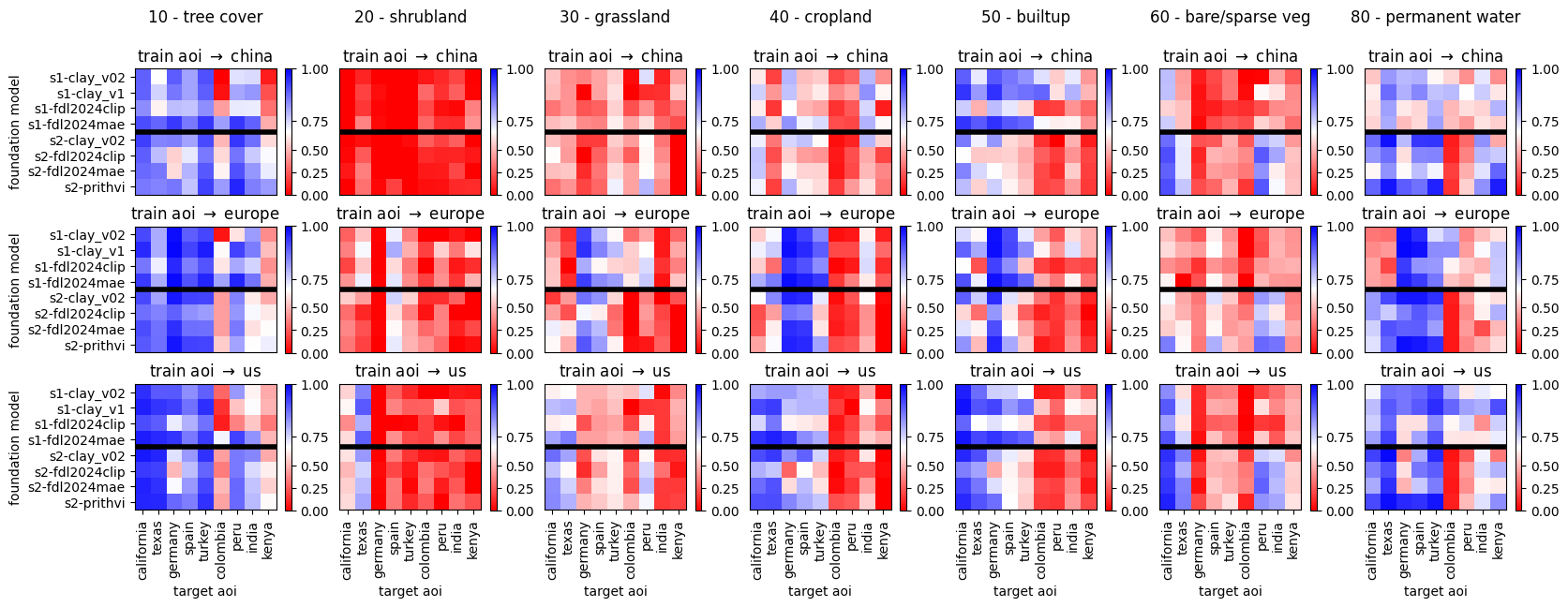}
\label{fig:fms-traintargets}
\end{figure}

Figure \ref{fig:fms-traintargets} shows an overall view of FM embeddings performance when trained on the different external AOIs and tested on our target AOIs. We can already observe several aspects. The US and Europe have reasonable generalizability across their regions in most tasks (observe, the first four columns on each chart corresponding to US on Spain and Germany, and Europe on California and Texas)

Tasks \texttt{tree cover} and \texttt{permanent water} show reasonable generalizability to non US and non European AOIs. Models for \texttt{permanent water} trained in Europe seem to work better in California and Texas only if we use Sentinel 2.

Some tasks seem very AOI specific. In task \texttt{cropland} we get some signal when transferring models between the US, Europe and Turkey. And also in models trained in Chine applied in India. We believe this might be due to the inherent differences between AOIs (different crops around the world) and these classes probably gather a large variety of land features (like crops of palm trees look very different from crops of soy). Task \texttt{shrubland} shows very low performance and no spatial generalizability at all and task \texttt{grassland} only timidly within the US and within Europe.

Other tasks seem inherently dependant on the sensor. For instance, generalizability in Perú and India for \texttt{bare/sparse vegetation} seems to occur when using Sentinel 2 FMs regardless where the models were trained, and \texttt{builtup} seems more generalizable with Sentinel 1 (bluer above the black horizontal line)x|x|, which is in line with the expected interaction of a SAR signal with man made objects.

\paragraph{Uncertainty in target AOIs when training with external data}

\begin{figure}[h]
\caption{Selected ablations increasing the number of elements for test chips in the target AOI used represented with \textbf{dot size} in the set of values [10,50,100,500]. \textbf{Squared markers} represent models with Sentinel-2 input, round ones with Sentinel-1 input. Thresholds of 0.7 correlation coefficient mean and 0.05 standard deviation are shown}
\centering
\includegraphics[width=1.0\textwidth]{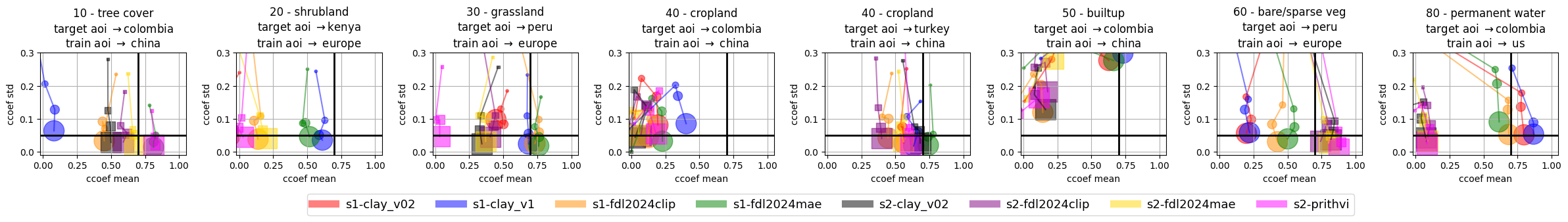}
\label{fig:selected-ablations}
\end{figure}

Figure \ref{fig:selected-ablations} shows some of the ablations on the number of elements used in target AOIs to test models trained on the external AOIs. Each chart corresponds to a column in Figure \ref{fig:fms-traintargets} and they were selected as they seem border cases seldom overcoming the 0.7 correlation coefficient threshold we established. Since we assume we are on a limited labeling budget on target AOIs, we are interested in having the minimum amount of labels without loosing overall predictive performance and, most importantly, with metric stability. Since we are repeating the same data and model configuration 20 times, resampling every time, we measure both the mean performance (correlation coefficient) and its standard deviation. Therefore, we also establish a threshold on the correlation coefficient standard deviation at 0.05. 

\begin{figure}[h]
\caption{Predictions for one run of cases in Table \ref{tab:resultsexternal}. Color shows the percentage of the \texttt{esawc} class, either the target or the prediction. }
\centering
\includegraphics[width=1.\textwidth]{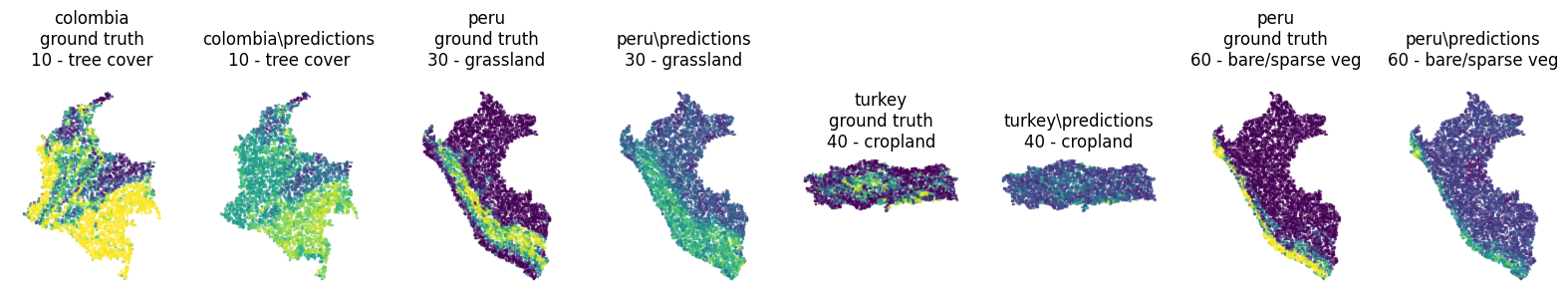}
\label{fig:countries}
\end{figure}

Then, Table \ref{tab:resultsexternal} details the cases we consider would be useful in practice with these thresholds (0.7 mean, 0.05 stdev). Observe that for some cases we are not able to get satisfactoyy FMs (although class \texttt{permanent water} for Colombia is left barely out with an standard deviation of 0.052). And for \texttt{builtup} in Colombia, even if we overcome the correlation coefficient threshold for its mean with 500 chips labeled, the uncertainty is quite high (around 0.3). Figure \ref{fig:countries} shows those predictions.

Finally, observe how the modality dependancy is much more explicit in several cases in Figure \ref{fig:selected-ablations} having well differentiated the performance of FMs using as input data S1 (circles) and S2 (squares).

\begin{table}[]
\begin{center}
\begin{tabular}{llllrrr}
\toprule
\specialcell{target\\aoi} & \specialcell{external\\aoi} & esawc class & FM & \specialcell{test\\elems} & \specialcell{corr coef\\ mean} & \specialcell{corr coef\\std} \\
\midrule
colombia & china & tree cover & s2-prithvi & 50 & 0.814 & 0.048 \\
peru & europe & grassland & s1-fdl2024mae & 100 & 0.764 & 0.045 \\
turkey & china & cropland & s1-fdl2024mae & 100 & 0.740 & 0.046 \\
peru & europe & bare/sparse veg & s2-prithvi & 100 & 0.894 & 0.030 \\
\bottomrule
\end{tabular}
\caption{Cases in the bottom right quadrants in Figure \ref{fig:selected-ablations} with the correlation coefficient mean is greater than 0.7, its standard deviation less than 0.05 and the number of test elements is lowest.
\label{tab:resultsexternal} 
}
\end{center}
\end{table}

\begin{figure}[h]
\caption{Same ablations as in Figure \ref{fig:selected-ablations} but using the same target AOI to split the data between train and validation, and using only the FM with which the best performance was obtained in Figure \ref{fig:selected-ablations}. \textbf{Dot size} represents the number of labeled chips for validation in the set [10,50,100,500]. \textbf{Shading} represents the number of labeled chips used for train within the target AOI in the set [10,50,100,500].}
\centering
\includegraphics[width=1.0\textwidth]{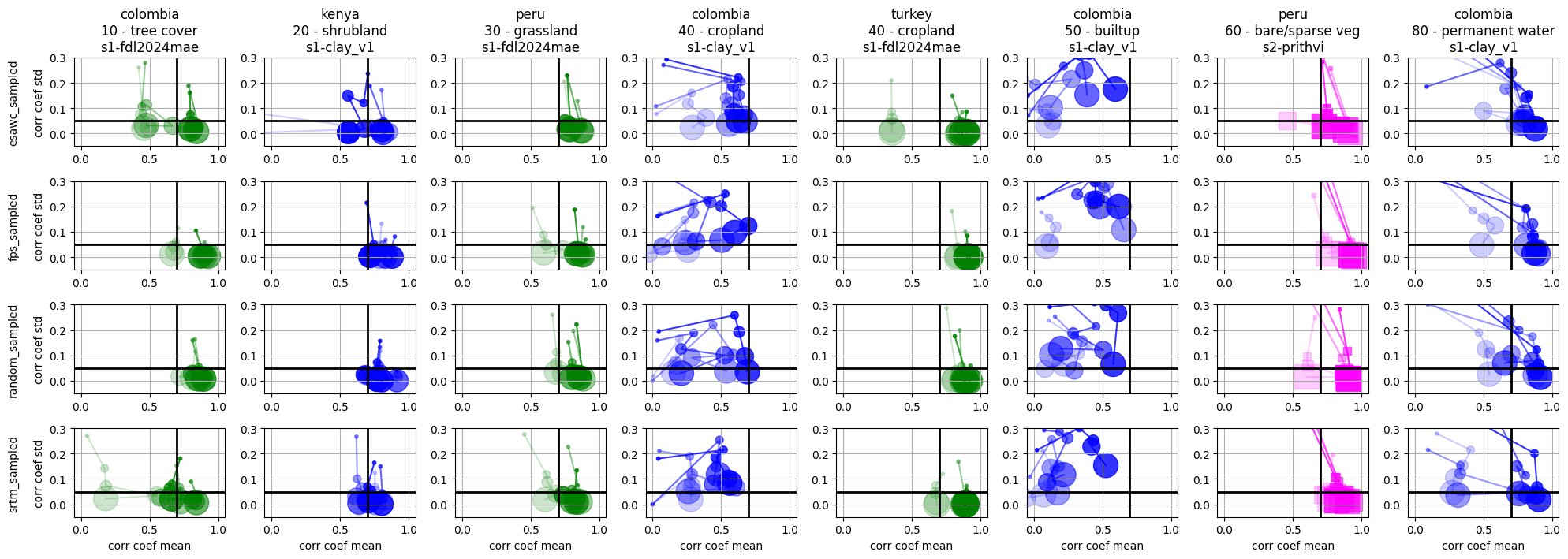}
\label{fig:selected-ablations-trainontarget}
\end{figure}

\paragraph{Uncertainty in target AOIs when training with its own data}
We now consider whether labeling budget on the target AOI can be split into train and validation and it is worthwhile as compared to using external AOIs data for training as shown above. Figure \ref{fig:selected-ablations-trainontarget} shows the correlation coefficient mean and standard deviation obtained as we ablate on the train and test split on each one of the selected target AOIs. We discriminate according to the sampling method used when training.

From those results, Table \ref{tab:resultstarget} shows the details of the cases where the correlation coefficient mean and standard deviation lies within the established thresholds. Observe first that, as could be expected, there is an overall increase in performance in two cases (\texttt{shrubland} in Kenya and \texttt{permanent water} in Colombia) and now we are within the performance thresholds. Observe that, in all cases, we can obtain high correlation coefficients at reduced uncertainty (standard deviation) with a handful of labeled chips split between train and test.

\begin{table}[h]
\begin{center}
\begin{tabular}{llllll}
\toprule
\specialcell{train aoi} & \specialcell{esawc class} & FM & \specialcell{number elems\\ \small{total (test/train)}} & sampling & corr coef \\
\midrule
colombia$^{\dagger}$ & tree cover & s1-fdl2024mae & 110 (10/100) & fps & 0.947 $\pm$0.032 \\
colombia$^{\ddagger}$ & tree cover & s1-fdl2024mae & 60 (50/10) & random & 0.816 $\pm$0.048 \\
\hdashline
kenya$^{\dagger}$ & shrubland & s1-clay\_v1 & 100 (50/50) & random & 0.923 $\pm$0.023 \\
kenya$^{\ddagger}$ & shrubland & s1-clay\_v1 & 60 (50/10) & fps & 0.791 $\pm$0.049 \\
\hdashline
peru$^{\dagger}$ & grassland & s1-fdl2024mae & 100 (50/50) & random & 0.888 $\pm$0.027 \\
peru$^{\ddagger}$ & grassland & s1-fdl2024mae & 60 (50/10) & esawc & 0.796 $\pm$0.050 \\
\hdashline
turkey$^{\dagger}$ & cropland & s1-fdl2024mae & 110 (10/100) & fps & 0.939 $\pm$0.035 \\
turkey$^{\ddagger}$ & cropland & s1-fdl2024mae & 60 (50/10) & fps & 0.836 $\pm$0.038 \\
\hdashline
peru$^{\dagger}$ & bare/sparse veg & s2-prithvi & 150 (50/100) & fps & 0.949 $\pm$0.032 \\
peru$^{\ddagger}$ & bare/sparse veg & s2-prithvi & 60 (50/10) & esawc & 0.896 $\pm$0.028 \\
\hdashline
colombia$^{\dagger}$ & permanent water & s1-clay\_v1 & 550 (500/50) & fps & 0.909 $\pm$0.035 \\
colombia$^{\ddagger}$ & permanent water & s1-clay\_v1 & 150 (100/50) & random & 0.905 $\pm$0.047 \\
\bottomrule
\end{tabular}
\caption{Cases in the bottom right quadrants for charts in Figure \ref{fig:selected-ablations-trainontarget} where the correlation coefficient mean is greater than 0.7 and its standard deviation less than 0.05. Cases with ${\ddagger}$ are the ones with the least number of total elements, and with $^{\dagger}$ are the ones where the correlation coefficient was greatest.
\label{tab:resultstarget} 
}
\end{center}
\end{table}

\paragraph{Sampling methods}

\begin{figure}[ht]
\caption{Effect of different sampling methods when we ablate on the number of train elements when using external AOIs for training models. Showing the mean of 20 repetitions for each configuration.}
\centering
\includegraphics[width=1.0\textwidth]{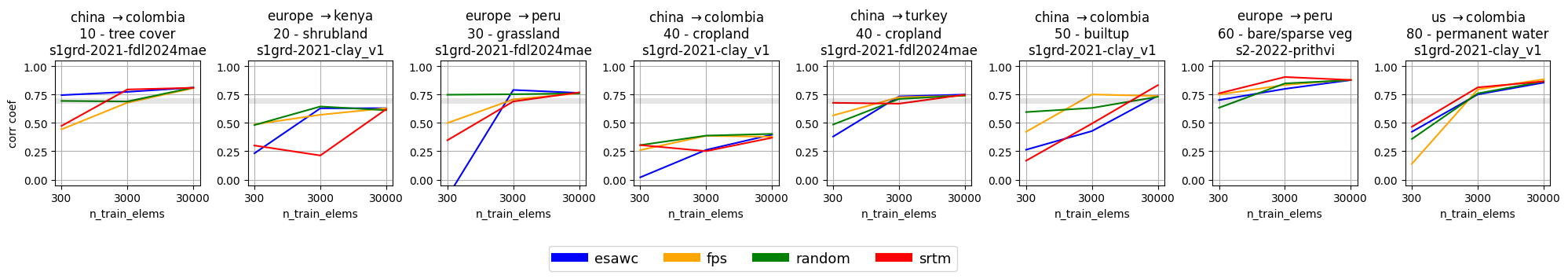}
\label{fig:externalaoi-sampling}
\end{figure}

We also make a reflection on the different sampling methods we used to select which data to use for training (whether on external or target AOIs). Observe how in Table \ref{tab:resultstarget} the sampling method becomes relevant when selecting the best performing cases. In particular Furthest Point Sampling (FPS) becomes a valuable alternative in the majority of the cases. Recall that FPS samples differently on each embeddings space, attempting to promote large euclidean distances within the embeddings of the sampled chips. These sampling differences also become particularly important when using external AOIs for training. Observe how in Figure \ref{fig:externalaoi-sampling}  with fewer elements (300 or 3000), the sampling method has great influence.

\section{Conclusion}
\label{sec:conclusion}
Through this study we have seen how generalizability and uncertainty vary greatly across different combinations of AOIs, tasks and input modalities. This shows that, when facing a labeling effort for a downstream task, it is key to make the right selection of FM and sampling method to make an efficient use of the labeling budget at hand.

As a result, we advocate for worldwide studies with \textbf{simple methods}, in which for each FM embeddings a number of \textbf{representative downstream tasks} globally available are tested against large combinatorics of train and target AOIs (countries, regions, continents, etc.), measuring generalizability and uncertainty with the methodology illustrated in this work. With proper experimentation such representative downstream tasks would be proxies of many other tasks related to them for which there are no global labels. Such studies would enable (1) better comparisons between FMs; and (2) pinpoint decisions on how to design downstream tasks within labeling budget constraints.

\begin{ack}
This work has been enabled by FDL Europe | Earth Systems Lab (https://fdleurope.org) a public / private partnership between the European Space Agency (ESA), Luxembourg Space Agency, Trillium Technologies, the University of Oxford in partnership with Google Cloud, NVIDIA Corporation, RSS Hydro, LuxProvide. We are thankful to the SAR-FM FDL 2024 Team and all the reviewers that participated in it.

\end{ack}

\bibliographystyle{unsrtnat}  
\bibliography{references} 


\end{document}